\documentclass[preprint,12pt]{elsarticle}

\usepackage{amssymb}
\usepackage{color}
\usepackage{amsmath,amsfonts}
\usepackage{booktabs}
\usepackage{graphicx}
\usepackage{multicol}
\usepackage{multirow}
\usepackage{colortbl}
\usepackage{subfigure}
\usepackage{cleveref}

\definecolor{color1}{RGB}{217, 85, 29}
\definecolor{color2}{RGB}{0, 114, 189}
\definecolor{color3}{gray}{0.9}

\journal{Pattern Recognition}

\usepackage{setspace}
\begin{document}

\begin{frontmatter}



\title{Visible-Infrared Person Re-Identification via Patch-Mixed Cross-Modality Learning}


\author[inst1]{Zhihao Qian}
\author[inst1]{Yutian Lin\corref{cor1}}

\affiliation[inst1]{organization={the School of Computer Science, Luojia laboratory, Wuhan University},
            city={Wuhan}, 
            postcode={430072}, 
            country={China}}

\author[inst1]{Bo Du}

\cortext[cor1]{Corresponding author}


\begin{abstract}
Visible-infrared person re-identification (VI-ReID) aims to retrieve images of the same pedestrian from different modalities, where the challenges lie in the significant modality discrepancy. 
To alleviate the modality gap, recent methods generate intermediate images by GANs, grayscaling, or mixup strategies. 
However, these methods could introduce extra data distribution, and the semantic correspondence between the two modalities is not well learned.
In this paper, we propose a Patch-Mixed Cross-Modality framework (PMCM), where two images of the same person from two modalities are split into patches and stitched into a new one for model learning. 
A part-alignment loss is introduced to regularize representation learning, and a patch-mixed modality learning loss is proposed to align between the modalities.
In this way, the model learns to recognize a person through patches of different styles, thereby the modality semantic correspondence can be inferred. 
In addition, with the flexible image generation strategy, the patch-mixed images freely adjust the ratio of different modality patches, which could further alleviate the modality imbalance problem.
On two VI-ReID datasets, we report new state-of-the-art performance with the proposed method.
\end{abstract}


\end{frontmatter}




\section{Introduction}

Visible-infrared person re-identification (VI-ReID)~\cite{ye2021deep} aims to match a target person between the RGB visible cameras and low-light infrared (IR) cameras. The task is increasing research interests~\cite{zahra2023person, gong2023spectrum, gavini2023thermal, wfcamrevit} because of its great value in the practical 24-hour surveillance system.
The main challenge of VI-ReID lies in modality discrepancy, where different wavelengths bring significantly different visual appearances (\textit{e.g.,} color, texture).

Typically, there are three main types of methods: 1) the representation learning based methods~\cite{huang2023exploring, zhang2024learning}, where networks are designed to learn discriminative features in a modality-shared space; 2) the metric learning based methods~\cite{MCSL,zhu2023visible}, where loss functions are designed to bridge the modality gap; 3) the modality-transform based methods~\cite{img, MMN}, which transform modalities into each other for style consistency. However, these methods try to handle the large modality discrepancy directly, while it is hard to align heterogeneous modalities without considering the correspondence between IR and RGB images.

\begin{figure}[ht]
    \centering
    \includegraphics[width=0.80\textwidth]{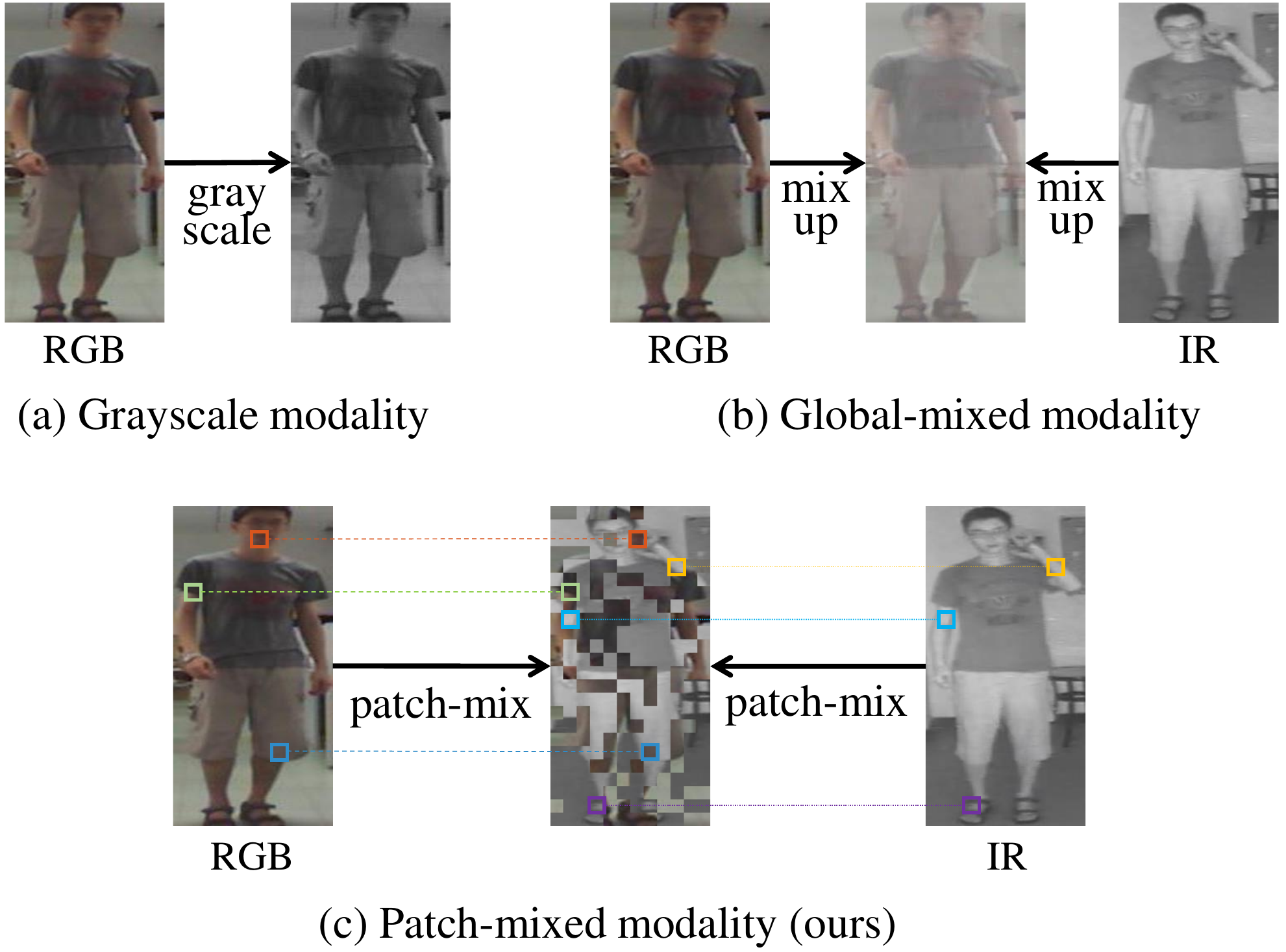}
    \caption{Different methods of generating the intermediate modality. (a) grayscale images are generated by visible images, (b) a mixed image is a global mixture of corresponding RGB and IR images, and (c) our proposed patch-mixed image, where each patch is either from RGB or IR, helps to infer the semantic corresponds between the two modalities and relieves the modality imbalance problem.}
    \label{figure1}
\end{figure}

To handle the above issue, recent works generate a third modality to assist cross-modality learning. Among them, a branch of works constructs the new modality upon only one modality. For example, \cite{X-modality} generates the third auxiliary modality by transforming the visible images to one-channel images and then reconstructs three-channel images. 
In~\cite{HAT}, grayscale images are generated by visible images and are utilized to enhance the robustness against color variations. However, as shown in Fig.~\ref{figure1} (a), these new images are generated through only one modality, where the information from the IR images is ignored. Another branch of works~\cite{MID, SMCL} adopts mixup~\cite{Mixup} strategy to generate intermediate modality images between RGB and infrared images, achieving promising results. However, as shown in Fig.~\ref{figure1} (b), it introduces a different data distribution, with each pixel is neither RGB nor IR, which may improve the generalization ability but the semantic correspondence between the two modalities is chaotic that the person may appear as a double shadow with two heads.

In this paper, we propose a Patch-Mixed Cross-Modality framework (PMCM), which leverages modality discrepancy by learning with a new patch-mixed modality.
As shown in Fig.~\ref{figure1} (c), the patch-mixed image is generated by blending a person's image of two modalities in the patch level where the data distribution within each patch is consistent with the original modality.
The patch-mixed modality benefits cross-modality learning in two aspects: 1) The semantic correspondence between two modalities could be inferred by recognizing a patch-mixed image. For example, the network will learn that the hair in the IR patch and the face in the RGB patch together describe a person's appearance. Thereby, the model learns to deal with the two modalities in the same way, and the modality gap is reduced. 
2) The patch-mixed modality also relieves the modality imbalance problem. As there usually are more images captured by daytime cameras, the data of IR and RGB images are imbalanced. With the flexible image generation approach, the proportion of different modalities can be adjusted freely to produce images with more IR information or RGB information. Therefore, the distribution of training samples is modified, which achieves an effect similar to over-sampling. 

As a minor contribution, we adopted an improved center-to-center loss to directly reduce the modality gap by aligning identity centers between RGB, IR and our patch-mixed modality. 
To regularize representation learning of part features, we take full advantage of the association between the part features and the global feature. Thus, a part alignment loss is proposed to constrain the consistency of part and global prediction distributions. With the part-based learning strategy, the discriminative part features are explored, which benefits the global feature learning in return. 
Besides, considering the shared information between the new modality and the other two modalities, we propose a patch-mixed modality learning loss to enhance the modality invariance learning by aligning the distribution of the prediction logits.

The main contributions of our work can be summarized as follows:
\begin{itemize}
    \item We propose a novel patch-mixed cross-modality learning framework for the VI-ReID task, which effectively encourages the model to treat the RGB and IR images in the same way and alleviate the modality imbalance problem. 
    \item We consider different constraints to further enhance the learned model.
    A part-alignment loss is proposed to constrain the consistency of part and global prediction distributions for more discriminative representation. 
    A patch-mixed modality learning loss is proposed to align the new modality with the other two modalities.
    \item Experimental results show that our method outperforms other methods on two VI-ReID datasets by a large margin, and the data imbalance problem is effectively alleviated.
\end{itemize}


\section{Related Work}

\subsection{Visible-Infrared Person Re-identification}
VI-ReID aims to match persons of different modalities, which faces the challenge of large intra-modality variation and inter-modality discrepancy. Wu~\textit{et al.}~\cite{zero-padding} was the first to define the task, which proposed a deep zero-padding method along with a large-scale VI-ReID dataset named SYSU-MM01. 

Following that, researchers propose to learn modality-specific and modality-shared feature representations by designing networks or loss functions. 
Ling~\textit{et al.}~\cite{MCSL} propose a Multi-Constraint similarity learning method that jointly considers the cross-modality relationships from three different aspects. 
Sun~\textit{et al.}~\cite{DCLNet} performs pixel-to-pixel dense alignment acting on the intermediate representations.
Huang~\textit{et al.}~\cite{huang2023exploring} makes use of both modality shared appearance features and modality-invariant relation features to boost performance
Huang~\textit{et al.}~\cite{mtl} take the initiative to investigate the importance and strategy of exploiting person body information.
Wan~\textit{et al.}~\cite{g2da} explicitly utilizes body topology to jointly achieve semantic- and structural-level alignment.

On the other hand, another branch of work aims to bridge the modality discrepancy by transforming the images from one modality to the other by generative adversarial networks (GANs). 
Choi~\textit{et al.}~\cite{Hi-CMD} propose an effective generator to extract pose-invariant and illumination-invariant features.
Zhao~\textit{et al.}~\cite{CICL} learns the color-irrelevant features and aligns the identity-level feature distributions. Zhang~\textit{et al.}~\cite{FMCNet} compensates for the missing modality-specific information from the other modality in the feature level.

\subsection{VI-ReID with Intermediate Modality Images}

To further reduce the modality gap, researchers propose to construct a third modality to assist shared feature space learning. 
In \cite{X-modality}, a third auxiliary modality is generated by transforming the visible images to one-channel images and then reconstructing three-channel images. 
In~\cite{HAT}, grayscale images are generated by visible images and are utilized to enhance the robustness against color variations. 
Following that, \cite{MMN} transforms both of the two modalities into the grayscale for modality alignment to reduce the modality discrepancy. However, in these methods, the third modality is generated upon only one modality, without considering the relationship between IR and RGB images.

Recently, some methods have proposed to generate intermediate modality images between RGB and infrared images, which achieved promising results.  
In~\cite{CMM}, inspired by mixup~\cite{Mixup}, a linear interpolation is performed of two images from different modalities for the same identities to generate mixed images. Similarly, in \cite{MID}, mixed modality images are generated with a dynamic mixup ratio learned by a deep reinforcement learning framework. In \cite{SMCL}, a syncretic modality collaborative learning model is designed, where shallow representation is mixed. 
Lu~\textit{et al.}~\cite{img} proposes an intermediate modality generation module involves CutMix~\cite{cutmix} to better integrate features from visible and infrared modalities.

Compared with these extra-modality learning methods, our Patch-Mix strategy generates the third modality by mixing the RGB and IR modalities at the raw pixel level, which helps to learn a semantic alignment of two modalities by a unified input.

\subsection{Modality Imbalanced Learning}
Current research on imbalanced data focuses on the class imbalance problem and introduces two main strategies: re-sampling and re-weighting. Re-sampling like \cite{cui2019class,huang2016learning} over-sample classes with few samples and under-sample classes with many samples. Re-weighting like \cite{cao2019learning,shen2016relay} adaptively adjusts the weights of different classes in the loss function. Liu~\textit{et al.}~\cite{liu2022learning} first notice the unique data imbalance problem in cross-modality tasks and name it Modality Imbalance, which refers to the situation that one modality contains more samples than the other modality. 
To address the problem, they borrow the idea of re-weighting and allowing independent augmentation for a specified modality. Different from their work, our PMCM alleviates this problem by adjusting the ratio of patch-mix, where more patches of one modality can be contained for data balance. 


\section{Proposed Method}
\begin{figure*}[t]
    \centering
    \includegraphics[width=\textwidth]{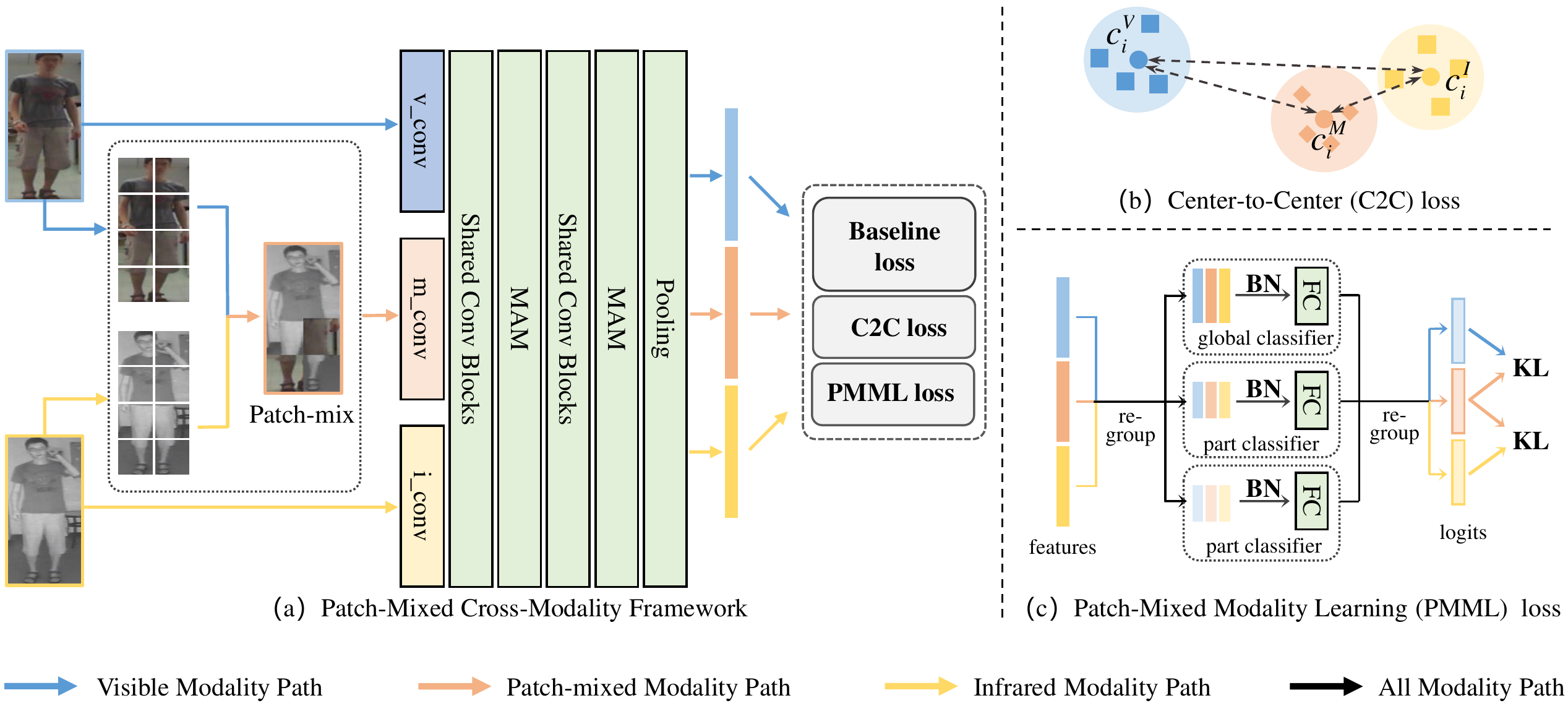}
    \caption{(a) Framework of the proposed PMCM. The patch-mixed image and the original images are together fed into the backbone network to extract features. After pooling, the obtained features are jointly optimized by the baseline loss, center-to-center loss, and patch-mixed modality learning loss. (b) Center-to-center (C2C) loss attempts to reduce the distance between the identity centers of any two modalities. (c) Patch-Mixed Modality Learning (PMML) loss aims to align the prediction distributions of the patch-mixed modality with that of the other two modalities, where global and local features are considered.
    }
    \label{figure2}
\end{figure*}

In this paper, we aim to learn modality-invariant representations by an intermediate patch-mixed modality, where cross-modality retrieval can be achieved. The overview of the proposed method is shown in Fig.~\ref{figure2}, where RGB, IR, and patch-mixed images are fed into the network, optimized by the baseline losses, center-to-center loss, and patch-mixed modality learning loss.

Following, we introduce our method in detail. We begin with the baseline method for this task. Then we illustrate the modality center alignment and patch-mixed cross-modality framework. 


\subsection{Baseline Method} 
\subsubsection{Baseline}
We adopt a two-stream network as our baseline and sample the same number of RGB and IR images in a mini-batch. We introduce the ResNet-50 as the backbone, where the first convolution blocks are modality-unique to learn modality-invariant low-level features, and the others are weight-shared to capture discriminative features. Besides, we alter channel-wise attention to part-wise attention in MAMs proposed by Wu~\textit{et al.}~\cite{MPANet} and insert them into the backbone to extract modality-irrelevant feature maps. After the backbone, we only adopt global average pooling to the obtained feature maps to attain global features and then use a batch normalization layer to get the query representations. At last, the loss function for the baseline is formulated as follows:

\begin{equation}
\mathcal{L}_{base\_global}=\mathcal{L}_{id,g}+\mathcal{L}_{tri}+\lambda_1\mathcal{L}_{s2s,g},
\end{equation}

\noindent where $\mathcal{L}_{id,g}$ is the cross-entropy loss of the global features after a classifier, $\mathcal{L}_{tri}$ is the hard triplet loss~\cite{triplet}, $\lambda_1$ denotes the weight of $\mathcal{L}_{s2s,g}$, which is the sample-to-sample loss~\cite{MMN}, attempting to pull close the sample features of different modalities with the same identity. Specifically, given the global features of two modalities, RGB and IR, the sample-to-sample loss is formulated as:

\begin{equation}
\mathcal{L}_{s2s}=\dfrac{1}{N}\sum\limits_{i=1}^{N}mean[F(f^{V}_{i})-F(f^{I}_i)],
\end{equation}

\noindent where $N$ is the number of paired samples in a mini-batch, $f_i^{V}$ and $f_i^{I}$ are the features of the i-th sample of RGB and IR, respectively. $F(\cdot)$ is a network with two fully-connected layers.

\subsubsection{Baseline with Part-based Learning} \label{mu}
Inspired by PCB \cite{PCB}, recent VI-ReID works~\cite{MMN, CICL, MPANet} learn part-based features to enhance global discriminative representation learning and achieve promising performance. In this work, we also exploit horizontal stripes to obtain part features. Similarly, cross-entropy loss and sample-to-sample loss are adopted to address the part-based features:
\begin{equation}
\mathcal{L}_{base\_part}=\mathcal{L}_{id,p}+\mathcal{L}_{s2s,p}.
\end{equation}
where $\mathcal{L}_{id,p}$ is the cross-entropy loss of all the part features and $\mathcal{L}_{s2s,p}$ is the sample-to-sample loss of all the part features.

In addition, since the global feature and part features both describe the same identity, we hope that the output distribution of part features could be similar to the global feature. Therefore, we calculate the KL divergence of the two output distributions as the regularization term, to learn more generalized part features. In this way, the prediction of global and each part features are supposed to be consistent. Given part and global features, the part alignment loss can be formulated as:

\begin{equation}
\begin{array}{ll}
\mathcal{L}_{part\_align} = \sum\limits_{i=1}^{N} \sum\limits_{k=1}^{P} {C_g(f^{g}_{i}) \log \dfrac{C_g(f^{g}_{i})}{C_{p_k}(f^{p_k}_{i})}} ,
\end{array}
\end{equation}

\noindent where P is the number of parts, $f^{p_k}_{i}$ denotes the feature of the k-th part feature of the i-th identity, $C_g(\cdot)$ and $C_{p_k}(\cdot)$ are the classifier of global features and that of the k-th part features.

Besides, in order to speed up the convergence of the model, we reduce the weight of losses involving part features in the early stage of training, which may cause relatively large interference. Therefore we set a weight $\mu$ to losses involving part features, which linearly increases with epochs and reaches its maximum at some point.
In this way, the total loss of the part-based baseline is formulated as follows:
\begin{equation}
\mathcal{L}_{base}=\mathcal{L}_{base\_global}+\mu(\mathcal{L}_{base\_part}+\mathcal{L}_{part\_align}).
\end{equation}

\subsection{Modality Center Alignment}

To further reduce the cross-modality variance, we consider directly aligning the identity centers between any two modalities. We adopt a global center-to-center loss similar to previous work \cite{MCSL}, where the distance between each center of the same identity from different modalities is minimized as shown in Fig.~\ref{figure2}(b). We obtain global centers of all training data by maintaining a memory bank instead of calculating the centers only in a mini-batch. Thus our global centers are more robust and tolerant to the mini-batch calculation bias caused by insufficient samples.

Given the global features of RGB images and IR images, the global center-to-center loss can be defined as:

\begin{equation}
\begin{array}{ll}
\mathcal{L}_{c2c,g}=\dfrac{1}{Y}\sum\limits_{i=1}^{Y}{\left\|m^V_i-m^{I}_i\right\|}^2,

\end{array}
\end{equation}

\noindent where $m^{V}_i$ and $m^{I}_i$ respectively denote the memory banks of the center of the $i$-th identity in the RGB and IR modality, which are updated every mini-batch, and $Y$ is the number of identities. 

Similarly, we consider the center relations between part features and use $\mu$ to balance the global and part losses. The final center-to-center loss is calculated as: 
\begin{equation}
\begin{array}{ll}
\mathcal{L}_{c2c}=\lambda_2\mathcal{L}_{c2c,g} + \mu \lambda_3 \mathcal{L}_{c2c,p}

\end{array}
\end{equation}
\noindent where $\lambda_2$ and $\lambda_3$ are the weights of $\mathcal{L}_{c2c,g}$ and $\mathcal{L}_{c2c,p}$.

In addition, there are also some differences in the optimization strategy. 
Since the stored features in the memory are inconsistent with those trained in the current training batch, 
we set a threshold epoch to delay the optimization of this loss until the model gets stable.

\subsection{Patch-Mixed Cross-Modality Learning}

To deal with the modality variance, recent VI-ReID works~\cite{HAT, MID, CMM} construct intermediate modalities and achieve promising performance. Different from these grayscale or mixup-based methods, we propose a novel patch-mixed intermediate modality, where each patch is from the original IR and RGB images, having the same data distribution of the original data.

\subsubsection{Patch-Mix strategy}

\begin{figure}
    \centering
    \includegraphics[width=0.9\textwidth]{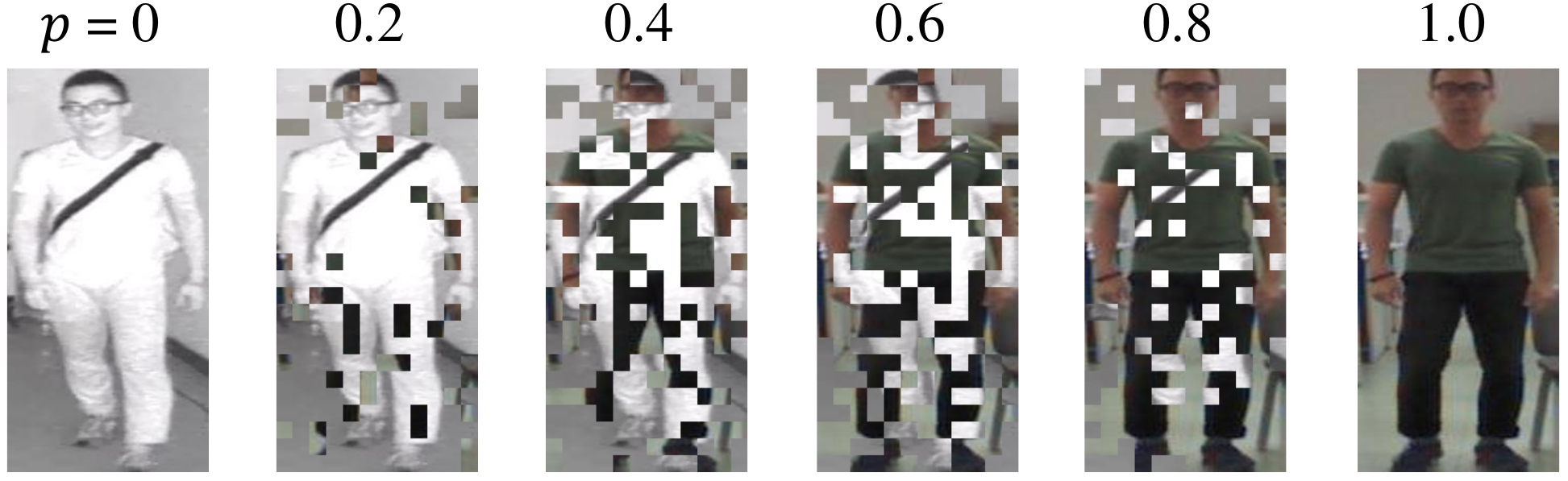}
    \caption{Patch-mixed images with different mix ratios $p$. When $p=0$, the image is composed of only an infrared image. As $p$ increases, more visible patches are adopted.}
    \label{figure3}
\end{figure}

Given an infrared image $x^I$ and a visible image $x^V$, a patch-mixed image $x^M$ is generated, where each patch $x^M(i,j)$ is formed by either $x^V(i,j)$ or $x^I(i,j)$. Here, $i$ and $j$ denote the patch index of image length and width, respectively. We set the probability of choosing an RGB patch to be $p$, and then the probability of choosing an infrared patch is $1-p$. As shown in Fig.~\ref{figure3}, the ratio $p$ balances the information from two modalities. 

The patch mixed images encourage cross-modality learning from two aspects. First, by learning from adjacent RGB and IR patches, which may represent the same semantic part, the model is encouraged to recognize the same semantic part through two modality patches. Therefore, the modality variance is bridged and the generalization ability is improved.
Second, the patch-mixed images help to reduce the modality imbalance. Usually, images captured by daytime cameras are more than those of nighttime cameras, so the data of IR and RGB images are typically imbalanced. As shown in Fig.~\ref{figure3}, with an adjustable ratio $p$, images with more IR information or RGB information can be generated freely according to the data distribution.

\subsubsection{Patch-Mixed Modality Learning (PMML)} Considering that the mixed modality contains internal information in both of the two modalities, we align it with the other two modalities. 

As shown in Fig.~\ref{figure2} (c), we input three training samples in three modalities with the same identity into the network and obtain their output distribution. The KL divergence of the two output distributions is calculated to constrain the learning of patch-mixed modality. Since the logits are grouped by global and part relationship after the classifiers, we first regroup the logits by modality and then align between the mixed modality and the two original modalities. 

Given part and global features, the part alignment loss can be formulated as:
\begin{equation}
\begin{array}{ll}
\mathcal{L}_{pmml}^{M,V} = \sum\limits_{i=1}^{K}{C_g(f^{V,g}_{i}) \log \dfrac{C_g(f^{V,g}_{i})}{C_{p_k}(f^{M,g}_{i})}} \\ 
+\sum\limits_{i=1}^{K}\sum\limits_{k=1}^{P}{C_{p_k}(f^{V,{p_k}}_{i}) \log \dfrac{C_{p_k}(f^{V,p_k}_{i})}{C_{p_k}(f^{M,{p_k}}_{i})}}
\end{array}
\end{equation}

\noindent where $f^{p_k}_{i}$ denotes the feature of the k-th part feature of the i-th identity, 
$C_g(\cdot)$ and $C_{p_k}(\cdot)$ are the classifier of global features and that of the k-th part features. 

Similarly, the alignment between the patch-mixed images and infrared images is calculated. The total patch-mixed modality learning loss is defined as:

\begin{equation}
\mathcal{L}_{pmml} = p\mathcal{L}_{pmml}^{M,V} + (1-p)\mathcal{L}_{pmml}^{M,I}.
\end{equation}

Note that to further alleviate the modality imbalance problem, we adopt a weight $p$ to balance the two losses, which is the same as the ratio of patch-mix. The effect of ratio $p$ is that the more image information of one modality the mixed modality contains, the more similar it will be to the source modality.

\subsection{Overall Optimization}

Ultimately, we optimize the PMCM in an end-to-end manner with the final loss defined as follows:

\begin{equation}
\begin{aligned}
\mathcal{L} = \mathcal{L}_{base} + \mathcal{L}_{c2c} + \mu\mathcal{L}_{pmml}
\end{aligned}
\end{equation}

\section{Experiments}

\subsection{Experimental Settings}

\subsubsection{Datasets} We evaluate our proposed framework on two VI-ReID datasets, SYSU-MM01 \cite{zero-padding} and RegDB \cite{regdb}.

The SYSU-MM01 dataset contains 491 identities captured by 4 visible cameras and 2 infrared cameras both including indoor and outdoor environments. The training set contains 22258 visible images and 11909 infrared images involving 395 identities, while the testing set contains 96 identities with 3803 infrared images as query images. Following the protocols, we test it both in all-search mode and indoor-search mode for only single-shot.

The RegDB dataset contains 412 identities with 206 identities for training and 206 identities for testing, where each identity has 10 visible images and 10 infrared images from a pair of overlapping visible and infrared cameras. Following the protocols, we test it both in Visible2Thermal mode, where visible images as query and infrared images as the gallery, and Thermal2Visible mode similar to the former.

\subsubsection{Evaluation metrics} 
For both datasets, we adopt the Cumulative Matching Characteristic (CMC) and mean Average Precision (mAP) to evaluate the performance and take the average result of ten tests to report.

\subsubsection{Implementation details} 
We implement our method with PyTorch and use ResNet50 pre-trained on ImageNet \cite{imagenet} as the backbone. All the input images are data augmented with a sequence of being resized to the size of $3 \times 384 \times 192$, random horizontal flipping, and random channel erasing \cite{cre}. The size of a mini-batch is set to 32, where we randomly sample 4 identities for each modality and 4 images for each identity. Besides, we adopt SGD as the optimizer with a weight decay of $5\times 10^{-4}$, a momentum of 0.9, and a dynamic learning rate schedule, where the rate linearly increases from 0 to 0.1 in the first 10 epochs and decreases by 0.1 per 30 epochs after the 30th epoch. The total number of training epochs is set to 101 and the number of tests is 10. In terms of hyper-parameters, we set $\lambda_1$, $\lambda_2$ and $\lambda_3$ to 0.2, 0.2 and 1.0. Following \cite{triplet}, we set the margin of triplet loss to 0.3. The ratio $p$ of patch-mix is set to 0.1 for SYSU-MM01 and 0.5 for RegDB. The parameter $\mu$ is gradually increased from 0 to 0.5 in the first 50 rounds and kept until the end.

We train our framework with one Tesla V100 GPU and the time cost of one epoch is increased by 54\% compared to that of baseline. However, our inference time is exactly the same as the baseline, which is more important in applications in the real-world scene.

\subsection{Comparison with State-of-the-Art Methods}

\begin{table*}[t]
    \resizebox{\textwidth}{!}{
    \centering
    \begin{tabular}{rcccccccccccccccc}
        \toprule
        \multirow{3}{*}{Method} & \multicolumn{8}{c}{All-search} & \multicolumn{8}{c}{Indoor-Search}  \\
        \cmidrule{2-17}
            &\multicolumn{4}{c}{Single-Shot} & \multicolumn{4}{c}{Multi-Shot} & \multicolumn{4}{c}{Single-Shot} & \multicolumn{4}{c}{Multi-Shot} \\
        \cmidrule{2-17}
        &R1 & R10 & R20 & mAP & R1 & R10 & R20 & mAP & R1 & R10 & R20 & mAP & R1 & R10 & R20 & mAP\\
        \midrule
        X-Modality~\cite{X-modality} & 49.92 & 89.79 & 95.96 & 50.73 & - & - & - & - & - & - & - & - & - & - & - & -\\
        MMN~\cite{MMN} & 70.6 & 96.2 & 99.0 & 66.9 & - & - & - & - & 76.2 & 97.2 & 99.3 & 79.6 & - & - & - & -\\
        MPANet~\cite{MPANet} & 70.58 & 96.21 & 98.80 & 68.24 & 75.58 & 97.91 & 99.43 & 62.91 & 76.74 & 98.21 & 99.57 & 80.95 & 84.22 & 99.66 & 99.96 & 75.11\\
        MCSL~\cite{MCSL} & 64.82 & - & - & 60.81 & 68.05 & - & - & 51.84 & - & - & - & - & - & - & - & - \\
        JCCL~\cite{CICL} & 57.20 & 94.30 & 98.40 & 59.30 & 60.70 & 95.20 & 98.60 & 52.60 & 66.60 & 98.80 & 99.70 & 74.70 & 73.80 & 99.40 & 99.90 & 68.30\\
        DCLNet~\cite{DCLNet} & 70.79 & - & - & 65.18 & - & - & - & - & 73.51 & - & - & 76.80 & - & - & - & - \\
        FMCNet~\cite{FMCNet} & 66.34 & - & - & 62.51 & 73.44 & - & - & 56.06 & 68.15 & - & - & 74.09 & 78.86 & - & - & 63.82\\
        CMT~\cite{CMT} & 71.88 & 96.45 & 98.87 & 68.57 & 80.23 & 97.91 & 99.53 & 63.13 & 76.98 & 97.68 & 99.64 & 79.91 & 84.87 & 99.41 & 99.97 & 74.11 \\
        MTL~\cite{mtl} & 67.25 & 95.38 & 98.46 & 64.29 & 72.95 & 96.94 & 99.27 & 57.62 & 69.58 & 96.66 & 99.03 & 74.37 & 80.39 & 98.80 & 99.83 & 68.60 \\
        MTMFE~\cite{huang2023exploring} & 69.47 & 96.42 & 99.11 & 66.41 & 73.74 & 97.48 & 99.59 & 59.78 & 71.72 & 97.19 & 98.97 & 76.38 & 81.49 & 99.18 & 99.79 & 70.94 \\
        $\mathrm{G}^2$DA~\cite{g2da} & 63.94 & 93.34 & 97.29 & 60.73 & 71.23 & 95.93 & 98.59 & 54.99 & 71.06 & 97.31 & 99.47 & 76.01 & 80.83 & 98.50 & 99.77 & 68.88 \\
        SIDA~\cite{gong2023spectrum} & 68.36 & 95.91 & 98.56 & 64.19 & - & - & - & - & 73.28 & 97.35 & 99.52 & 77.49 & - & - & - & - \\
        LDAEF~\cite{zhang2024learning} & 66.61 & - & - & 62.86 & 75.24 & - & - & 56.70 & 70.90 & - & - & 75.78 & 81.84 & - & - & 69.42 \\

        PMT~\cite{pmt} & 67.53 & 95.36 & 98.64 & 64.98 & - & - & - & - & 71.66 & 96.73 & 99.25 & 76.52 & - & - & - & - \\
        MRCN-P~\cite{mrcn} & 70.8 & 96.5 & 99.1 & 67.3 & - & - & - & - & 76.4 & 98.5 & 99.9 & 80.0 & - & - & - & - \\
        ProtoHPE~\cite{protohpe} & 71.92 & 96.19 & 97.98 & 70.59 & - & - & - & - & 77.81 & 98.64 & 99.59 & 81.31 & - & - & - & - \\

        \midrule
        CMM~\cite{CMM} & 51.80 & 92.72 & 97.71 & 51.21 & 56.27 & 94.08 & 98.12 & 43.39 & 54.98 & 94.38 & 99.41 & 63.7 & 60.42 & 96.88 & 99.5 & 53.52\\
        SMCL~\cite{SMCL} & 67.39 & 92.87 & 96.76 & 61.78 & 72.15 & 90.66 & 94.32 & 54.93 & 68.84 & 96.55 & 98.77 & 75.56 & 79.57 & 95.33 & 98.00 & 66.57\\
        MID~\cite{MID} & 60.27 & 92.90 & - & 59.40 & - & - & - & - & 64.86 & 96.12 & - & 70.12 & - & - & - & -\\
        IMG~\cite{img} & 61.31 & 91.31 & & 57.20 & 69.79 & 95.12 & & 51.01 & 67.20 & 96.06 & & 72.41 & 78.14 & 97.69 & & 65.51 \\
        \rowcolor{color3} PMCM(ours) & \textbf{75.54} & \textbf{97.49} & \textbf{99.30} & \textbf{71.16} 
        & \textbf{82.52} & \textbf{99.00} & \textbf{99.78} & \textbf{65.88} 
        & \textbf{81.52} & \textbf{98.99} & \textbf{99.71} & \textbf{84.33} 
        & \textbf{90.06} & \textbf{99.80} & \textbf{99.97} & \textbf{79.45} \\
        \bottomrule
    \end{tabular}}
    \caption{Comparison of CMC and mAP performances with the SOTAs on SYSU-MM01. Particularly, methods in the last five lines adopt different mixup strategies to generate an intermediate modality for model learning. 
    }
    \label{table1}
\end{table*}

\begin{table}[tp]
    \centering
    \resizebox{0.5\textwidth}{!}{
    \begin{tabular}{rcccc}
        \toprule
        \multirow{2}{*}{Method} & \multicolumn{2}{c}{Visible2Infrared} & \multicolumn{2}{c}{Infrared2Visible}  \\
        \cmidrule{2-5}
        & Rank-1 & mAP & Rank-1 & mAP\\
        \midrule
        X-Modality~\cite{X-modality} & 62.20 & 60.20 & - & -\\
        MMN~\cite{MMN} & 91.6 & 84.1 & 87.5 & 80.5\\
        MPANet~\cite{MPANet} & 82.8 & 80.7 & 83.7 & 80.9\\
        MCSL~\cite{MCSL} & 93.83 & 87.55 & 91.55 & 85.25 \\
        JCCL~\cite{CICL} & 78.8 & 69.4 & 77.9 & 69.4\\
        DCLNet~\cite{DCLNet} & 81.2 & 74.3 & 78.0 & 70.6\\
        FMCNet~\cite{FMCNet} & 89.12 & 84.43 & 88.38 & 83.86\\
        CMT~\cite{CMT} & \textbf{95.17} & 87.3 & \textbf{91.97} & 84.46 \\

        MTL~\cite{mtl} & 89.91 & 85.64 & 88.34 & 84.06 \\
        MTMFE~\cite{huang2023exploring} & 85.04 & 82.52 & 81.11 & 79.59 \\
        $\mathrm{G}^2$DA~\cite{g2da} & 73.95 & 65.49 & 69.67 & 61.98 \\

        SIDA~\cite{gong2023spectrum} & 81.73 & 75.07 & 79.71 & 72.60 \\
        LDAEF~\cite{zhang2024learning} & 90.76 & 87.30 & 88.79 & 85.44 \\
        PMT~\cite{pmt} & 84.83 & 76.55 & 84.16 & 75.13 \\
        MRCN-P~\cite{mrcn} & 95.1 & 89.2 & 92.6 & 86.5 \\
        ProtoHPE~\cite{protohpe} & 88.74 & 83.72 & 88.69 & 81.99 \\

        \midrule
        CMM~\cite{CMM} & 59.81 & 60.86 & - & -\\
        SMCL~\cite{SMCL} & 83.93 & 79.83 & 83.05 & 78.57\\
        MID~\cite{MID} & 87.45 & 84.85 & 84.29 & 81.41\\
        IMG~\cite{img} & 89.70 & 85.82 & 87.64 & 84.03 \\
        \midrule
        \rowcolor{color3} PMCM(ours) & 93.09 & \textbf{89.57} & 91.44 & \textbf{87.15} \\
        \bottomrule
    \end{tabular}}
    \caption{Comparison of CMC and mAP performances with the SOTAs on RegDB.}
    \label{table2}
\end{table}

We compare the proposed PMCM with several state-of-the-art methods for VI-ReID, including 
X-Modality~\cite{X-modality}, MMN~\cite{MMN}, MPANet~\cite{MPANet}, SMCL~\cite{SMCL}, MCSL~\cite{MCSL}, JCCL~\cite{CICL}, DCLNet~\cite{DCLNet}, FMCNet~\cite{FMCNet}, MID~\cite{MID}, CMT~\cite{CMT}, SIDA~\cite{gong2023spectrum}, LDAEF~\cite{zhang2024learning}, PMT~\cite{pmt}, MRCN-P~\cite{mrcn}, ProtoHPE~\cite{protohpe}, MTL~\cite{mtl}, MTMFE~\cite{huang2023exploring}, $\mathrm{G}^2$DA~\cite{g2da}, IMG~\cite{img}, and CMM~\cite{CMM}.

The comparison results on SYSU-MM01 and RegDB are respectively shown in Table~\ref{table1} and Table~\ref{table2}. We observe that our PMCM outperforms the existing SOTAs on all evaluation metrics by a large margin in SYSU-MM01. PMCM is superior to the second-best ProtoHPE in single-shot and all-search mode in SYSU-MM01 by 3.62\% in Rank-1 accuracy and 0.57\% in mAP. Compared with the methods adopting mixup schemes, PMCM also shows the best, exceeding SMCL in single-shot and all-search mode in SYSU-MM01 by 8.15\% in Rank-1 accuracy and 9.38\% in mAP, demonstrating that our methods can encourage the model to learn the semantic correspondence between the two different modalities thus improving the overall performance.

Although some methods like MSCL, CMT, and MRCN-P surpass us in certain metrics in RegDB, their performance in SYSU-MM01 is significantly inferior to our PMCM. For instance, CMT exceeds our PMCM by 2.08\% in Rank-1 accuracy in Visible2Infrared mode and 0.53\% in Rank-1 accuracy in Infrared2Visible mode, but CMT is inferior to our PMCM by 3.66\% in Rank-1 accuracy and 2.59\% in mAP in single-shot and all-search mode in SYSU-MM01. Compared with them, PMCM exhibits a better overall performance, providing evidence for effectively alleviating the modality imbalanced problem.

All the results above fully demonstrate the superiority and robustness of our PMCM, where more modality-invariant and discriminative features can be learned.

\subsection{Algorithm Analysis}

\subsubsection{Ablation studies}


\begin{table}[tp]
    \centering
    \setlength{\tabcolsep}{3pt}
    \begin{tabular}{lc|ccccc|cc}
        \toprule
        &B           & Part       & PartAlign  & C2C        & PatchMix   & PMML       & Rank-1 & mAP   \\
        \midrule
        1&\checkmark  & -          & -          & -          & -          & -          & 64.58 & 62.41 \\
        2&\checkmark  & \checkmark & -          & -          & -          & -          & 67.24 & 64.90 \\
        3&\checkmark  & \checkmark & \checkmark & -          & -          & -          & 69.52 & 65.37 \\
        4&\checkmark  & -          & -          & \checkmark & -          & -          & 68.63 & 64.43 \\
        5&\checkmark  & \checkmark & \checkmark & \checkmark & -          & -          & 70.66 & 66.06 \\
        6&\checkmark  & -          & -          & -          & \checkmark & -          & 67.92 & 63.77 \\
        7&\checkmark  & -          & -          & -          & \checkmark & \checkmark & 69.08 & 64.20 \\
        8&\checkmark  & \checkmark & \checkmark & \checkmark & \checkmark & -          & 73.76 & 69.88 \\
        9&\checkmark  & \checkmark & \checkmark & \checkmark & \checkmark & \checkmark & 75.54 & 71.16 \\
        \bottomrule
    \end{tabular}
    \caption{Ablation studies on the effectiveness of each component of the proposed PMCM in SYSU-MM01, where B denotes the baseline method.}
    \label{table3}
\end{table}

To validate each component of PMCM, we conduct ablation experiments on SYSU-MM01 in the all-search and single-shot mode in an accumulation way. The experimental result is shown in Table~\ref{table3} and numbered by row.

\textbf{Effectiveness of the part-based baseline (Part).} When exploring part features upon the global-based baseline, the performance is increased by 2.66\% and 2.49\% on Rank-1 and mAP respectively, showing its enhancement to the feature representation learning.

\textbf{Effectiveness of the part alignment loss (PartAlign).} The part alignment loss plays the role of regularization to the global and part feature predictions. Comparing row 3 with row 2 in the table, we observe that the improvements in Rank-1 accuracy and mAP are 2.32\% and 0.47\%, proving the part alignment loss better mines discriminative part features.

\textbf{Effectiveness of the center-to-center loss (C2C).} Compared with the baseline, C2C improves the Rank-1 accuracy and mAP by 4.05\% and 2.02\% and collaboration with "Part" and "PartAlign" further improves the two metrics by 2.03\% and 1.63\%, thanks to that C2C pulls close the centers of different modalities of the same identity to extract modality-invariant features.

\textbf{Effectiveness of training with patch-mixed images (PatchMix).} Adding PatchMix strategy to the baseline increases the Rank-1 accuracy and mAP by 3.34\% and 1.36\%, respectively, illustrating its effective. After introducing the modules above, the PatchMix strategy can still provide a significant improvement of the two metrics by 3.10\% and 3.82\%.

\textbf{Effectiveness of patch-mixed cross-modality learning (PMML).} As an auxiliary learning scheme, PMML shows improvement of the Rank-1 accuracy and mAP in two further experiments based on PatchMix, demonstrating its ability to enhance the effectiveness of PatchMix.

\subsubsection{Analysis of patch-mix ratio}

\begin{figure}
    \centering
    \subfigure[SYSU-MM01]{
        \includegraphics[width=0.45\textwidth]{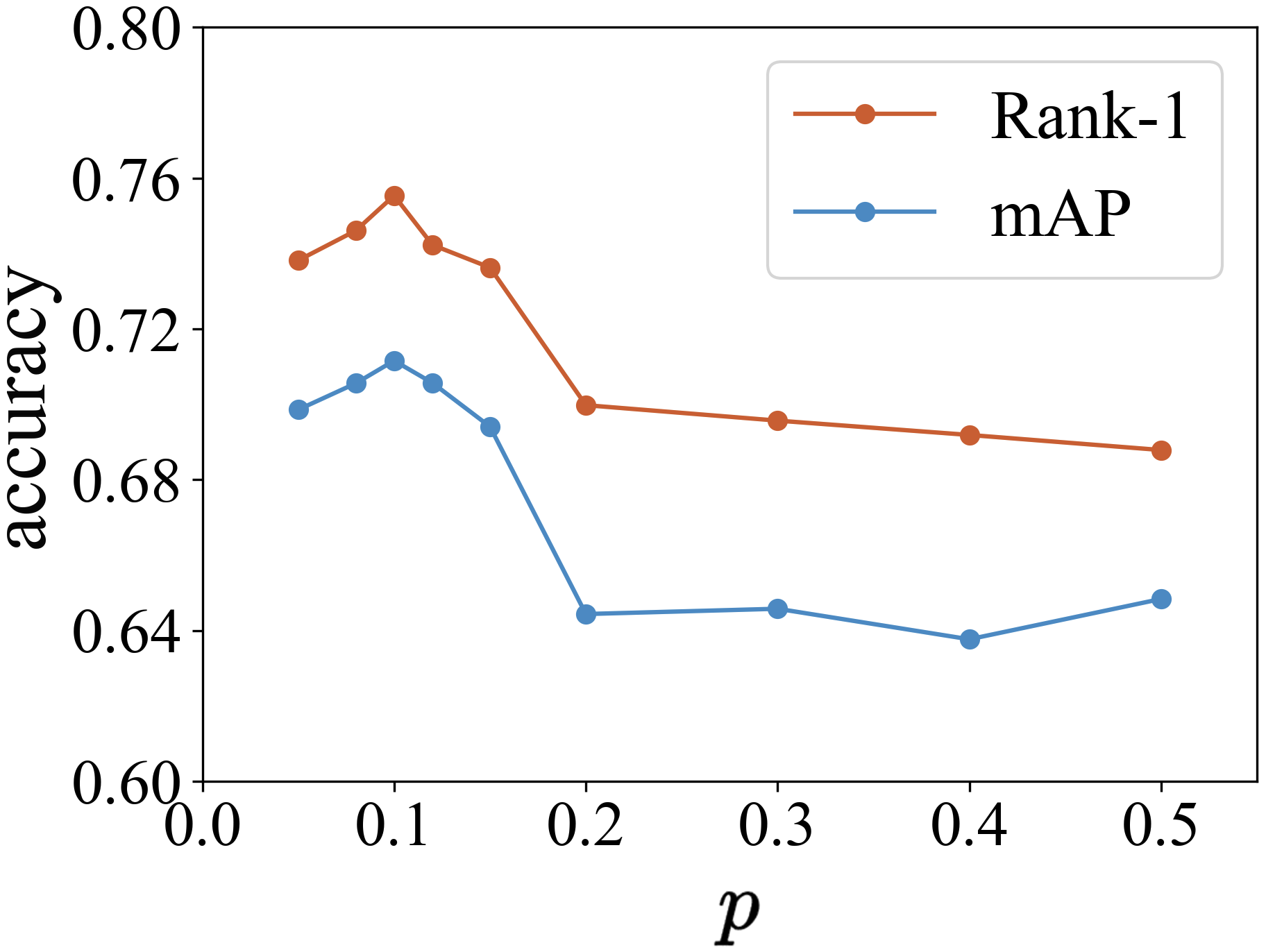}
    }\subfigure[RegDB]{
        \includegraphics[width=0.45\textwidth]{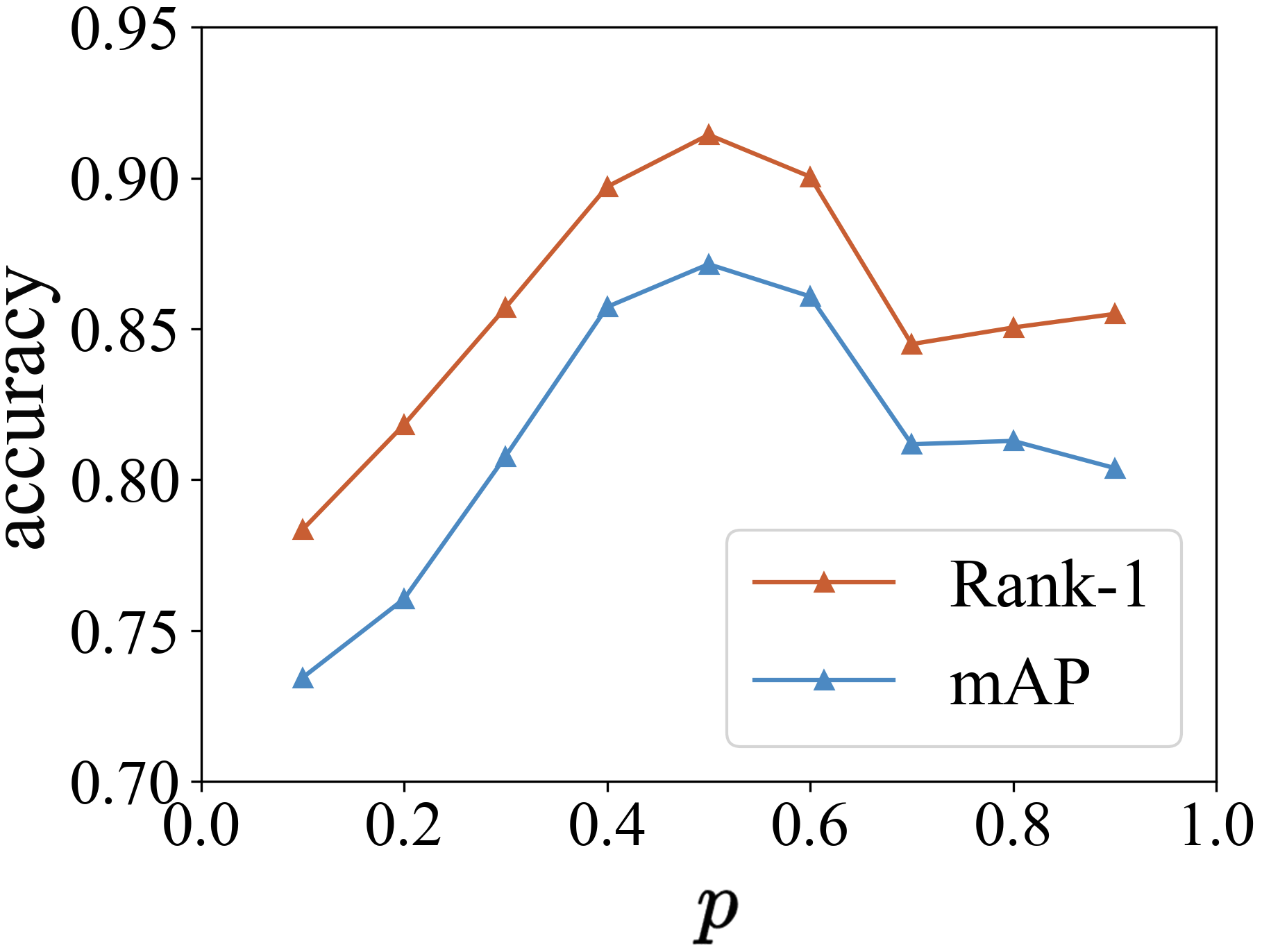}
    }
    \caption{Influence of the different values of patch-mix ratio $p$, (a) experiments on SYSU-MM01 in all-search and single-shot mode, and (b) experiments on RegDB in Infrared2Visible mode.}
    \label{figure4}
\end{figure}

The ratio $p$ adjusts the proportion of IR patches and RGB patches in the patch-mixed images. When $p$ is 0, the image is composed of only an IR image. As $p$ increases, more RGB patches are contained. 
Based on this, we argue that the model will ultimately learn $(1+p)$ times the RGB information and $(2-p)$ times the IR modality information. When the modality imbalanced problem occurs, we can rebalance the RGB and IR information by adjusting the value of $p$.
As shown in Fig.~\ref{figure4}, we evaluate the ratio $p$ on both SYSU-MM01 in all-search single-shot mode and RegDB in Infrared2Visible mode.

On SYSU-MM01, the samples of IR modality are much fewer than those of RGB modality. We observe that when the ratio $p$ is set to 0.5 (the number of patches of two modalities is equal), a relatively low re-ID performance is obtained. When $p$ decreases from 0.5 to 0.1 (more IR patches contained), the performance is continuously improved, and when the ratio is set to 0.1, the best performance is achieved. This demonstrates that, when there is more IR information contained in the intermediate modality, a better balance of IR and RGB data is kept, and the model could learn the two modalities equally. 

On RegDB, IR and RGB modalities have the same number of samples, which means RegDB is a data-balanced dataset. As shown in Fig.~\ref{figure4} (b), the model achieves best performance when $p$ is set to 0.5. In addition, when the ratio is larger or smaller than 0.5, the balance between the two modalities is broken, thus leading to a performance decrease. 

All pieces of evidence above prove that the proposed patch-mix scheme effectively alleviates the modality imbalance problem.

\subsubsection{Analysis of parameters $\lambda_1$, $\lambda_2$ and $\lambda_3$}

\begin{figure}[t]
    \centering
    \includegraphics[width=0.76\textwidth]{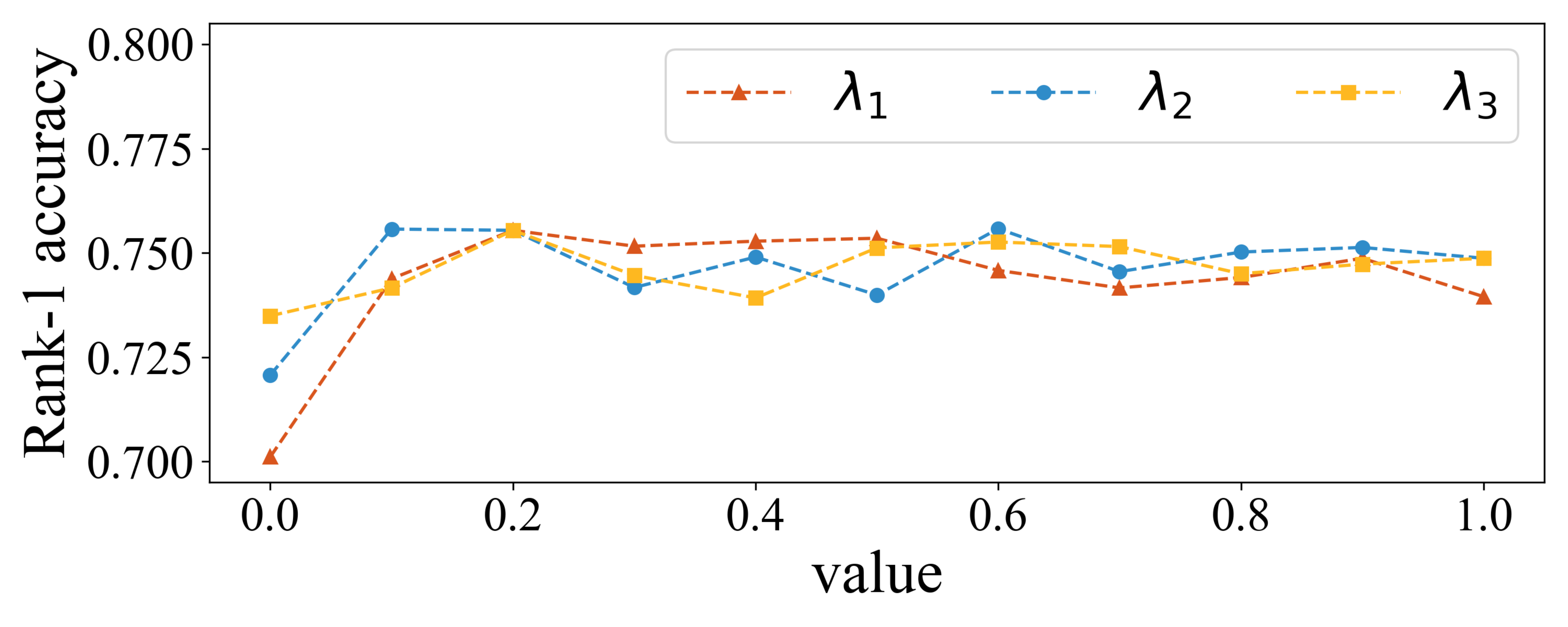}
    \caption{Analysis of the Rank-1 accuracy with parameters $\lambda_1$, $\lambda_2$ and $\lambda_3$. We keep the other parameters constant while testing the target one, and the table shows the insensitivity of our PMCM to these parameters.}
    \label{figure5}
\end{figure}

We have evaluated the parameters including $\lambda_1$, $\lambda_2$, and $\lambda_3$ by control variates in SYSU-MM01. The line chart of the rank-1 accuracy is shown in Fig.~\ref{figure5}, which reveals that the fluctuation range of the metrics is approximately 3\% within the range (0.1, 1.0) and if the value falls to 0, the accuracy will significantly decrease. The experiments demonstrate that our method is insensitive to the value of three loss weights but each of them counts.

\begin{table}[t]
    \centering
    \resizebox{0.7\textwidth}{!}{
    \begin{tabular}{cccc}
        \toprule
        Method & Rank-1 & Rank-10  & mAP \\
        \midrule
        CutMix~\cite{cutmix} & 69.48 & 95.07 & 64.73 \\
        RandomErasing~\cite{re} & 69.71 & 95.31 & 63.95 \\
        Grayscale & 70.94   & 95.95      & 65.84 \\
        PatchMix & \textbf{73.76}   & \textbf{97.34}  & \textbf{69.88} \\
    \hline
        Mixup~\cite{Mixup} + PMML     & 73.26   & 96.95      & 69.39 \\
        RandomErasing~\cite{re} + PMML & 73.74 & 97.02 & 70.00 \\
        CutMix~\cite{cutmix} + PMML & 73.90 & 97.16 & 70.31 \\
        PatchMix + PMML & \textbf{75.54} & \textbf{97.49}  &\textbf{71.16} \\
        \bottomrule
    \end{tabular}}
    \caption{Comparison of different intermediate modality generation strategies on SYSU-MM01.}
    \label{table4}
\end{table}

\subsubsection{Analysis of different mixup strategies}

In addition, to show the superiority of our method over other intermediate modality generation strategies, we compare images generated by grayscale, the standard mixup with PMML, CutMix~\cite{cutmix}, CutMix with PMML, RandomErasing~\cite{re} and RandomErasing with PMML. 
During implementation, we replace the PatchMix and PMML modules in our framework with the other generative methods and conduct experiments on them in the SYSU-MM01 dataset. 
The results are shown in Table~\ref{table4}. We observe that our method exceeds all of the strategies, which fully demonstrates the advantage of our patch-mix scheme over other existing methods. 
Moreover, our PMML scheme works quite effectively on the other intermediate modality generation strategies and produces different degrees of improvement in the two metrics.




\begin{table}[t]
   \centering
    \resizebox{0.6\textwidth}{!}{
   \begin{tabular}{ccccc}
       \toprule
       method          & Rank-1  & Rank-10 & Rank-20 & mAP \\
       \midrule
       Schedule1       & 71.89   & 96.38   & 98.63   & 67.14 \\
       Schedule2       & 73.91   & 97.19   & 98.94   & 69.57 \\
       ours            & 75.54   & 97.49   & 99.30   & 71.16 \\
       \bottomrule
   \end{tabular}}
   \caption{Analysis of the update schedule. Schedule1 is constant at 0.5, and Schedule2 is a linearly increasing value.}
   \label{table6}
\end{table}

\begin{table}[tp]
    \centering
    \resizebox{0.6\textwidth}{!}{
    \begin{tabular}{ccccc}
        \toprule
        Size & Rank-1 & Rank-10 & Rank-20 & mAP \\
        \midrule
        $4\times4$   & 74.47 & 97.38 & 99.30 & 69.92 \\
        $8\times8$   & 75.09 & 97.27 & 99.26 & 70.93 \\
        $12\times12$ & 75.13 & 97.31 & 99.27 & 70.86 \\
        $16\times16$ & \textbf{75.54} & \textbf{97.49} & \textbf{99.30} & \textbf{71.16} \\
        $32\times32$ & 74.62 & 97.13 & 99.11 & 69.92 \\
        \bottomrule
    \end{tabular}}
    \caption{Influence of different sizes of patch experimented on SYSU-MM01.}
    \label{table7}
\end{table}

\subsubsection{Analysis of the global-local balancing weight $\mu$. }

In order to speed up the convergence and reduce the error caused by the low-quality features extracted from part features in the early training stage, we design the special update schedule for the hyper-parameter, which linearly increases from 0 to 0.5 in the first half of the training phase and remains unchanged till the end. To validate the effectiveness of our schedule, we compare it with two other schedules. Schedule1 is constant at 0.5, and Schedule2 is a linearly increasing value represented by $current\_epoch / total\_epochs$. Based on the results shown in Table~\ref{table6}, our schedule achieves the best performance, which exceeds Schedule1 by 3.65\% and Schedule2 by 1.63\% in Rank-1 accuracy, which meets our expectations.

\subsubsection{Analysis of different sizes of the patch}

We conduct experiments to investigate the impact of different sizes of the image patches in SYSU-MM01.
According to the results shown in Table\ref{table7}, either too large or too small sizes will slightly reduce the effect, which means our method is quite robust to the patch size. Finally, we choose the patch size of $16\times16$ for our method.



\begin{figure*}[tp]
    \centering
    \includegraphics[width=0.98\textwidth]{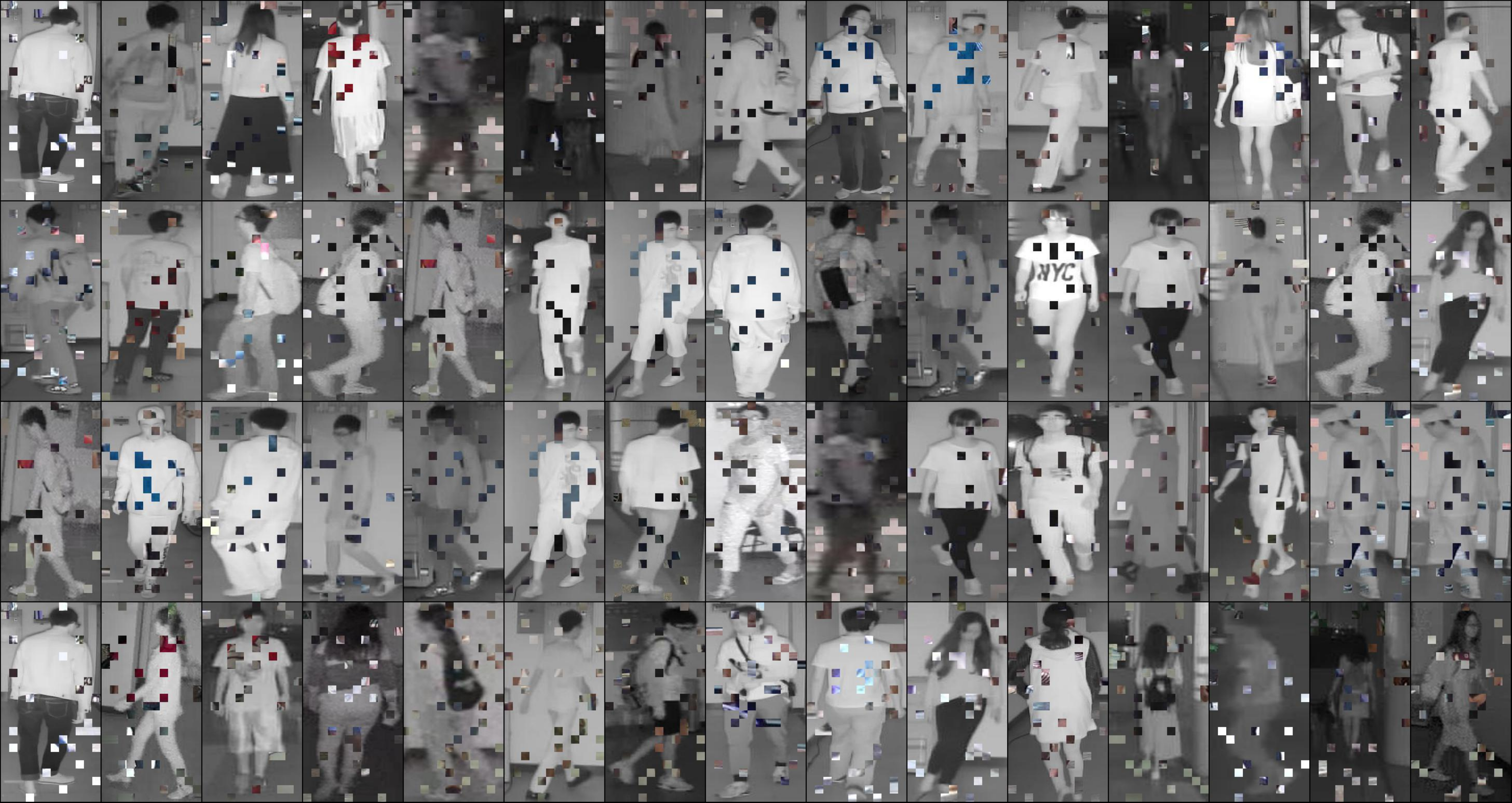}
    \caption{Visualization of PatchMixed images in SYSU-MM01 with the mixup ratio 0.1. It is obvious that all the generated images retain most of the infrared modality information and introduce a small amount of RGB modal information. Only very few of them have semantic misalignment issues, such as missing a piece on the face and replacing it with a piece from the surroundings.}
    \label{figure6}
\end{figure*}

\subsection{Visualization}

\subsubsection{Visualization of PatchMixed images}
We manually checked 200 random PatchMixed images in SYSU-MM01 with the mixup ratio of 0.1 and part of the PatchMixed images are shown in Fig.~\ref{figure6}. 
As a result, we find that in most generated images, the mixed patches are harmonious with the patches around them, while less than 10\% percent of generated mixed images have body part miss alignment between modalities. In addition, most misalignment is contour shifts within a body part. The case of containing two right hands (or two other body parts) has never happened. Although the edges of patches may not be always consistent, most of the time, the semantic meaning of a patch is consistent with its neighbors. 

\begin{figure}[!tp]
    \centering
    \subfigure[Baseline]{
        \includegraphics[width=0.44\textwidth]{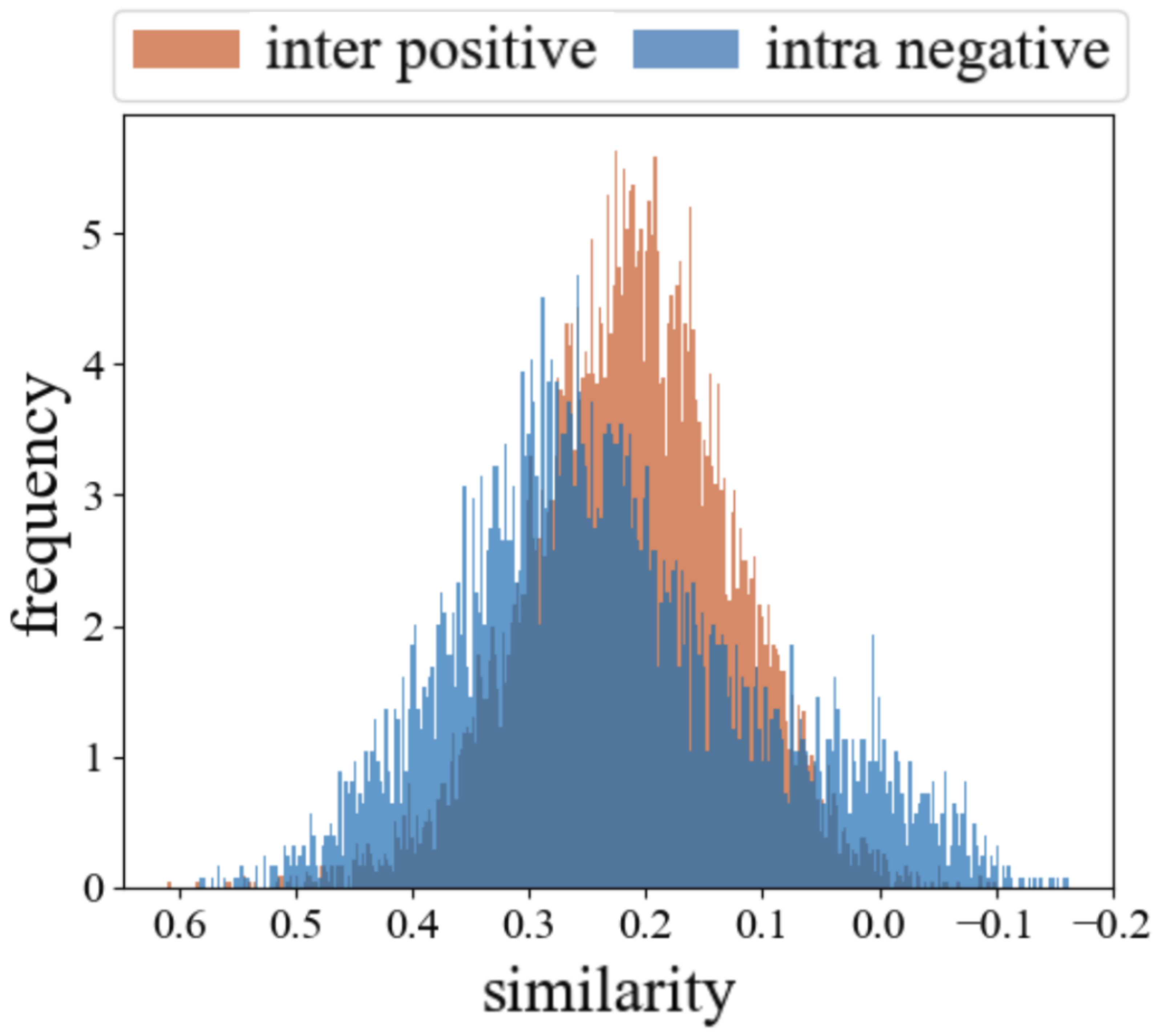}
        \label{figure7a}
    }\subfigure[PMCM]{
        \includegraphics[width=0.44\textwidth]{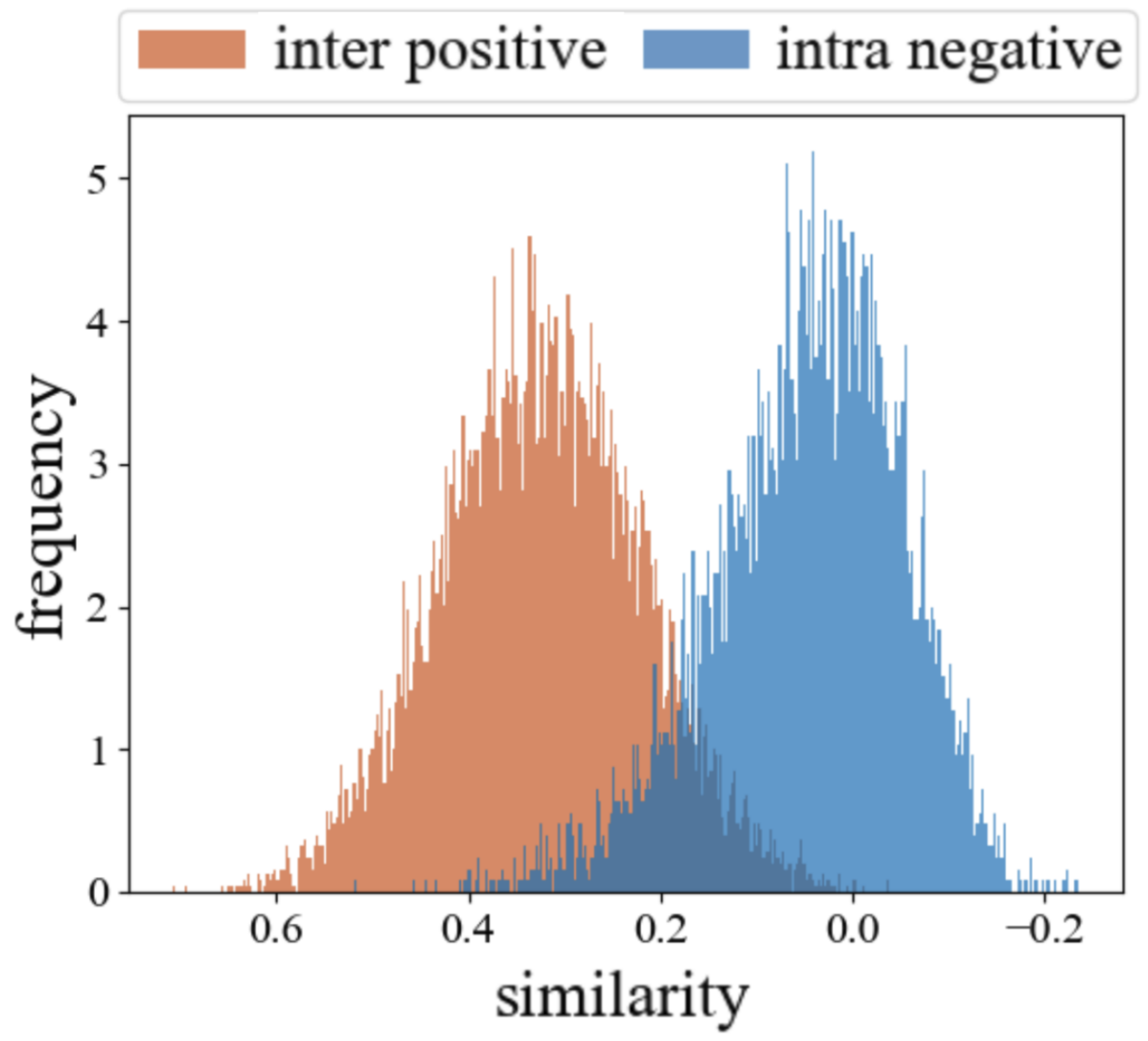}
        \label{figure7b}
    }
    \caption{Visualization of cosine similarity distribution of inter-modality positive samples and intra-modality negative samples with (a) baseline and (b) our PMCM on the test set of SYSU-MM01.}
    \label{figure7}
\end{figure}

\subsubsection{Cosine distance distribution}
We visualize the cosine distance distribution of inter-modality positive samples and intra-modality negative samples in the test set of SYSU-MM01, as is shown in Fig.~\ref{figure7}. In baseline (Fig.~\ref{figure7a}), the two types of image pair share a similar distance distribution, which reveals that the baseline can hardly retrieve inter-modality positive images. Instead, our proposed PMCM (Fig.~\ref{figure7b}) separates the inter-positive pairs and intra-negative pairs, showing that cross-modality images could have bigger similarity and be retrieved successfully. It fully demonstrates that our method can effectively reduce the intra-class modality discrepancy. 

\begin{figure}[!tp]
    \centering
    \subfigure[Baseline]{
        \includegraphics[width=0.4\textwidth]{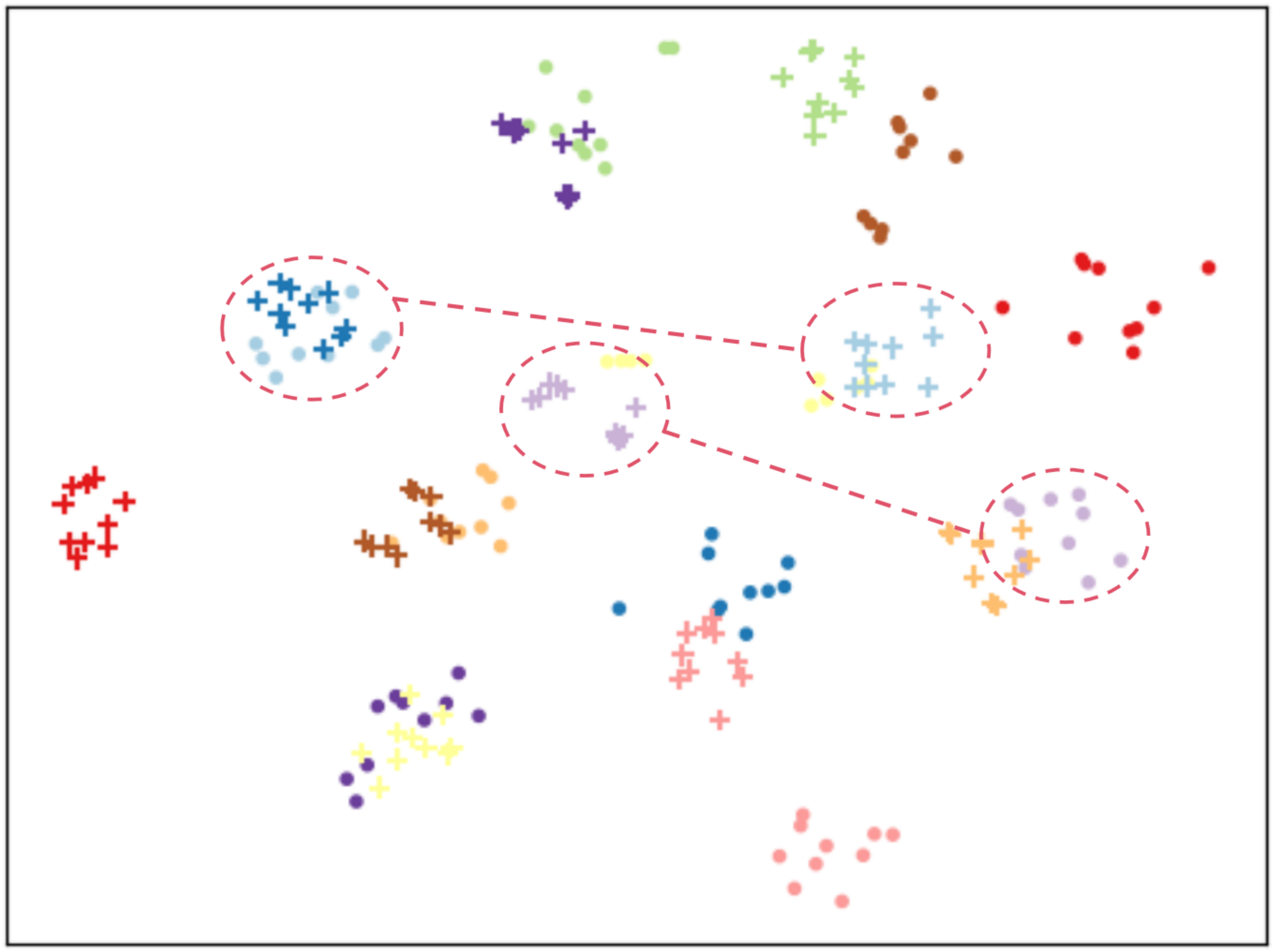}
        \label{figure8a}
    }\subfigure[PMCM]{
        \includegraphics[width=0.4\textwidth]{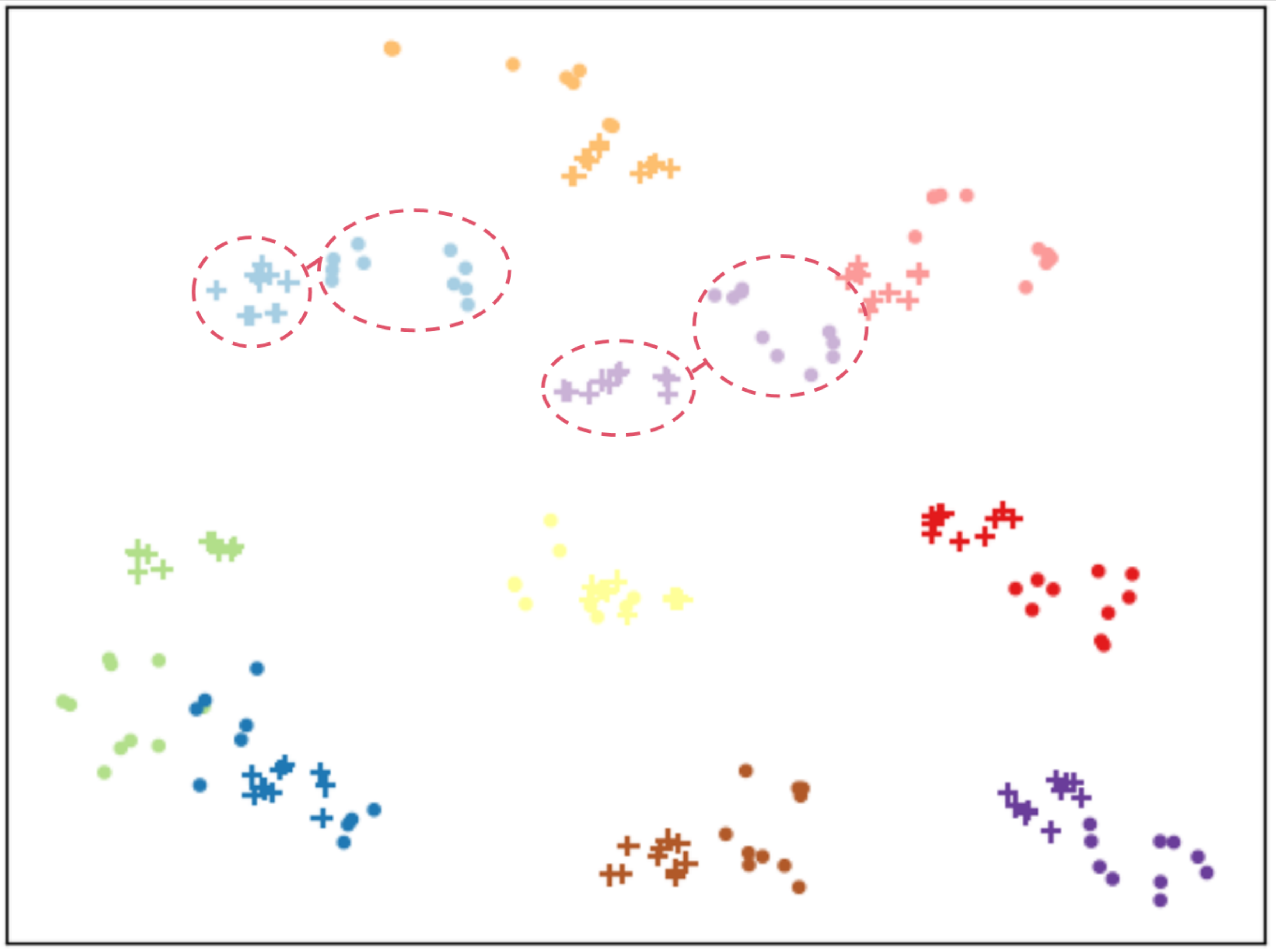}
        \label{figure8b}
    }
    \caption{Visualization of the feature embeddings distribution of (a) baseline and (b) PMCM via t-SNE in the SYSU-MM01 dataset, where dots and pluses in the same color denote features of the same identity in RGB modality and infrared modality.}
    \label{figure8}
\end{figure}

\subsubsection{Feature distribution}
To further explore the reason why PMCM is effective, we visualize the feature distribution of samples in the SYSU-MM01 dataset via t-SNE as shown in Fig.~\ref{figure8}, where different colors denote different identities. For the baseline (Fig.~\ref{figure8a}), feature embeddings of the same identity are far away, which indicates the baseline can hardly narrow the large modality gap. In comparison, our PMCM (Fig.~\ref{figure8b}) discriminates and aggregates these feature embeddings of the same identity separately and clearly, demonstrating that our method can effectively improve representation learning and narrow the cross-modality gap.


\section{Conclusion}
In this paper, we propose a Patch-Mixed Cross-Modality (PMCM) framework for VI-ReID. 
A patch-mixed modality is introduced to learn the semantic correspondence between visible and infrared images and alleviate the modality imbalance problem. A patch-mixed modality learning loss is adopted to take advantage of the third modality to reduce the modality gap between the two modalities. The patch-mixed modality and its learning strategy can be assumed as an add-on component and easily adopted in future work. 
To further reduce the inter- and intra-modality variance, we propose a part-alignment loss to constrain the consistency of part and global prediction distributions for more discriminative representation.
Extensive experiment results have demonstrated the superior effectiveness of PMCM compared with other state-of-the-art methods. 
The code will be publicly available, which will enable future research.
The limitation of our work is that some generated images may have body parts that miss alignment between modalities, which could lead to inaccurate representation learning.
In our future work, we plan to explore solutions to improve alignment in patch-mixed images.

 \bibliographystyle{elsarticle-num} 
 \bibliography{main}





\end{document}